\documentclass[journal]{IEEEtran}
\ifCLASSINFOpdf
\else
\fi

\usepackage{graphicx}

\usepackage{booktabs} 
\usepackage{float}
\usepackage{subcaption}
\usepackage{amsmath,amssymb,hyperref}
\usepackage{xcolor}
\usepackage{float}
 \usepackage[linesnumbered,boxed]{algorithm2e}

\usepackage{amsmath}
\usepackage{amssymb}
\usepackage{indentfirst}
\usepackage{cases}

\begin{document}
%
\title{Improved Design of Quadratic Discriminant Analysis
Classifier in Imbalanced Settings}
%
%
%

\author{Amine~Bejaoui,~\IEEEmembership{Student Member,~IEEE,}
        Khalil~Elkhalil,~\IEEEmembership{Member,~IEEE,}
		Abla~Kammoun, Mohamed-Slim~Alouini 
        and~Tareq~Al-Naffouri,~\IEEEmembership{Fellow,~IEEE}
\thanks{
A. Bejaoui, A. Kammoun, M.-S. Alouini and T. Alnaffouri are with the Electrical Engineering Program, King Abdullah University of Science and Technology, Thuwal, Saudi Arabia; e-mails: \{amine.bejaoui, abla.kammoun, slim.alouini, tareq.alnaffouri\}@kaust.edu.sa} 
\thanks{K. Elkahlil is with the Electrical Engineering Program, Duke University, Durham, North Carolina, USA; e-mail: \{khalil.elkhalil\}@duke.edu}}
\maketitle

\begin{abstract}
The use of quadratic discriminant analysis (QDA) or its regularized version (R-QDA) for classification is often not recommended, due to its well-acknowledged high sensitivity  to the estimation noise of the covariance matrix. This becomes all the more the case in imbalanced settings in which training data for each class are disproportionate and for which it has been  found that R-QDA becomes equivalent to the  classifier that assigns all observations to the same class. In this paper, we propose an improved R-QDA that is based on the use of two regularization parameters and a modified bias, properly chosen to avoid inappropriate behaviors  of R-QDA in imbalanced settings and to ensure the best possible classification performance. The design of the proposed classifier builds on a random matrix theory based analysis of its performance when the number of samples and that of features grow large simultaneously. 
	The performance of the proposed classifier is assessed on both real and synthetic data sets and was shown to be much better than what one would expect from a traditional R-QDA.
\end{abstract}


%
\IEEEpeerreviewmaketitle

\section{Introduction}
%
%
%
%
\IEEEPARstart{D}{iscriminant}  analysis encompasses a wide variety of techniques used for classification  purposes. These techniques, commonly recognized among  the class of model-based methods in the field of machine learning \cite{12}, rely merely on the fact that we assume a parametric model in which the outcome is described by a set of explanatory variables that follow a certain distribution. Among them, we particularly distinguish linear discriminant analysis (LDA) and quadratic discriminant analysis (QDA) as the most representatives. LDA is often connected or confused with Fisher discriminant analysis (FDA) \cite{fisher}, a method of projecting the data into a subspace and turns out to coincide with LDA when the target subspace has two dimensions. Both LDA and QDA are obtained by maximizing the posterior probability under the assumption that observations follow normal distribution, with the single difference that LDA assumes common covariances across classes while QDA assumes the most general situation with classes possessing different means and covariances.
If the data follow perfectly the normal distributions and the statistics are perfectly known, QDA turns out to be the optimal classifier that achieves the lowest possible classification error rate \cite{14}. It coincides with LDA when the covariances are equal but outperforms it when they  are different.  However, in practical scenarios, the use of LDA and to a large extent QDA was not always shown to yield the expected performances. This is because the mean and covariance of each class, which are in general unknown,  are  estimated based on available training data with perfectly known classes. The obtained estimates are then used as plug-in estimators in the classification rules associated with LDA and QDA.  The estimation error of the class statistics causes a provably degradation of the performances which reaches very high levels when the number of samples is comparable or less than their dimensions. In this latter situation, QDA and LDA, relying on computing the inverse of the covariance matrix could not be used. To overcome this issue, one technique consists in using a regularized estimate of the covariance matrix as a plug-in estimator of the covariance matrix giving the name to Regularized LDA (R-LDA) or Regularized QDA (R-QDA) to the associated classifiers. 
However, this solution does not allow for a significant reduction of the estimation noise. The situation is even worse for  R-QDA, since the number of samples used to estimate the covariance matrix of each class is lower than that of LDA. Moreover, in imblanced settings, the estimation quality of the covariance matrix associated with each class is not the same, one class possessing more samples than the other classes. These are probably the reasons why LDA provided in many scenarios better performances than QDA, although it might wrongly consider the covariances across classes  equal. 

A  question of major theoretical and practical interest is to investigate to which extent the estimation noise of the covariance matrix impacts the performances of R-LDA and R-QDA. In this respect, the study of LDA and subsequently that of R-LDA have received a particular attention, dating back to the early works of Raudys \cite{Raudys}, before being investigated again  using recent advances of random matrix theory tools in a recent series of works \cite{zollanvari,wang}. However, the theoretical analysis of QDA and R-QDA is more scarce and very often limited to specific situations in which the number of samples is higher than that of the dimensions of the statistics \cite{qda_stoachastic1}, or under specific structures of the covariance matrices \cite{8,9,10}. It was only recently that our work in \cite{kh} considered the analysis of R-QDA for general structures of the covariance matrices and identified the necessary asymptotic conditions under which QDA does not behave trivially by returing always the same class. Among these conditions is to assume that training data are balanced across classes. Indeed, as will be discussed later in this work,    
in case of imbalanced settings,   the difference in estimation quality of covariance matrices across classes make the classification rule of R-QDA keep asymptotically the same sign irrespective of the class of the testing observation. As a result, the use of the traditional discrimination rule of R-QDA is equivalent to assigning all observations to the same class.     

This lies behind the main motivation of the present work. Based on a careful investigation of the asymptotic behavior of R-QDA under imbalanced settings in binary classification problems, we propose a modified classification rule for R-QDA that copes with cases in which the proportions of training data from both classes are not equal. The new classification rule is based on using two different regularization parameters instead of a common regularization parameter as well as an optimized bias properly chosen to minimize the misclassification error rates. Interestingly, we show that the proposed classifier not only outperforms R-LDA and R-QDA but also other state-of-the-art classification methods, opening promising avenues for the use of the proposed classifier in practical scenarios.    

The rest of the paper is organized as follows:  
In section \ref{section:RQDA}, we provide an overview of the quadratic discriminant classifier and identify the issues related to the use of this classifier in imbalanced settings. In section \ref{section:improved_qda}, we propose an improved version of the R-QDA classifier that overcomes all  these problems and we design a consistent estimator of the misclassification error rate that can be used to properly chose the parameters of the proposed R-QDA and constitutes a valuable alternative to the traditional cross-validation approach. Finally, Section \ref{section:results} presents the results of a set of numerical simulations on both synthetic and real data that confirm our theoretical findings.

\noindent{\it Notations}
 {Scalars, vectors and matrices are respectively denoted by non-boldface, boldface lowercase and boldface uppercase characters. $\mathbf{0}_{p \times n}$ and $\mathbf{1}_{p \times n}$ are respectively the matrix of zeros and ones of size $p \times n$, $\mathbf{I}_{p}$ denotes the $p \times p$ identity matrix. The notation $\|.\|$ stands for the Euclidean norm for vectors and the spectral norm for matrices. $(.)^{ T}$ , ${\rm Tr}[.]$ and $|.|$ stands for the transpose, the trace and the determinant of a matrix respectively. For two functions f and g, we say that $f = O (g)$, if $\exists \  0 < M < \infty$ such that $|f | \leq M g$. Moreover, for $X$ random variable, $X=O_p(1)$ refers to a variable that is bounded in probability.   
We say also that that $f = \Theta (g)$, if $\exists 0 < C_1 < C_2 < \infty$ such that $C_1 g\leq|f | \leq C_2 g$. Moreover, we denote by
$ \stackrel{p}{\rightarrow} 0 $ and $\stackrel{as}{\rightarrow}$    the convergence in probability and the almost sure convergence of random variables. Finally $\Phi(.)$ denotes the cumulative density function (CDF) of the standard normal distribution, i.e. $\Phi(x)= \int_{-\infty}^{x} \frac{1}{\sqrt{2\pi}}e^{-\frac{t^{2}}{2} }dt$.}

\section{Regularized quadratic discriminant analysis}
\label{section:RQDA}
The asymptotic analysis carried out in \cite{kh} has made it clear that in case R-QDA is designed based on imbalanced training samples, it would asymptotically assign all testing observations to the same class. 
Such a behavior has led the authors in \cite{kh} to consider the analysis of R-QDA only under a balanced training sample. Interestingly, understanding such a behavior can be made through simple arguments based on  a close examination of the mean and variance of the classification rule associated with R-QDA.  These arguments do not necessitate random matrix theory results, thus, we find it important to present them at the outset in order to pave the way towards our improved classifier. 
But prior to that, let us first review  the traditional R-QDA for binary classification. 

\subsection{Regularized QDA for binary classification}
For ease of presentation, we focus on binary classification problems where we have two distinct classes. We assume that the data follow a Gaussian mixture model, such that observations in class $\mathcal{C}_i$, $i\in\{0,1\}$ are drawn from a multivariate Gaussian distribution with mean  $\boldsymbol{\mu}_{i}$ and covariance $\mathbf{\Sigma}_{i}$. More formally, we assume that  
\begin{align}
\mathbf{x}\in\mathcal{C}_i \ \ \Leftrightarrow \ \ \mathbf{x}=\boldsymbol{\mu}_{i}+\mathbf{\Sigma}_{i}^{1 / 2} \boldsymbol{z}, \quad \textnormal{with} \ \ \boldsymbol{z} \sim \mathcal{N}\left(\mathbf{0}, \mathbf{I}_{p}\right)
    \label{1}
\end{align}
Let  $\pi_{i}$, i = 0, 1 denote the prior probability that $\mathbf{x}$ belongs to class $\mathcal{C}_{i}$.
The classification rule associated with the QDA classifier  is given by
\begin{align}
  &W^{Q D A}(\mathbf{x})\nonumber \\&=-\frac{1}{2} \log \frac{\left|\boldsymbol{\Sigma}_{0}\right|}{\left|\boldsymbol{\Sigma}_{1}\right|}-\frac{1}{2} \mathbf{x}^{T}\left(\boldsymbol{\Sigma}_{0}^{-1}-\boldsymbol{\Sigma}_{1}^{-1}\right) \mathbf{x}+\mathbf{x}^{T} \boldsymbol{\Sigma}_{0}^{-1} \boldsymbol{\mu}_{0}\nonumber\\&-\mathbf{x}^{T} \boldsymbol{\Sigma}_{1}^{-1} \boldsymbol{\mu}_{1}  -\frac{1}{2} \boldsymbol{\mu}_{0}^{T} \mathbf{\Sigma}_{0}^{-1} \boldsymbol{\mu}_{0} +\frac{1}{2} \boldsymbol{\mu}_{1}^{T} \boldsymbol{\Sigma}_{1}^{-1} \boldsymbol{\mu}_{1}-\log \frac{\pi_{1}}{\pi_{0}}\label{eq:rule} 
\end{align}
which is used to classify the observations based on the following rule:
\begin{align}\label{3}
    \left\{ \begin{array}{ll}{\mathbf{x} \in \mathcal{C}_{0}}, & {\text { if } \quad W^{Q D A}>0} \\ {\mathbf{x} \in \mathcal{C}_{1}}, & {\text { otherwise. }}\end{array}\right.    
\end{align}
As seen from \eqref{eq:rule}, the classification rule of QDA involves the true parameters of the Gaussian distribution, namely the means and covariances associated with each class. In practice, these parameters are not known. One approach to solve this issue is to estimate them using the available training data. The obtained estimates are then used as plug-in estimators in \eqref{eq:rule}. 
In particular, consider the case in which $n_{i}, i \in\{0,1\}$ training observations for  each class $\mathcal{C}_{i}, i \in \{0,1\}$ are available and  denote by $\mathcal{T}_{0}=\left\{\mathrm{x}_{l} \in \mathcal{C}_{0}\right\}_{l=1}^{n_{0}}$  and $\mathcal{T}_{1}=\left\{\mathrm{x}_{l} \in \mathcal{C}_{1}\right\}_{l=n_{0}+1}^{n_{0}+n_{1}=n}$  their respective samples. The sample estimates of the mean and covariances of each class are then given by:
\begin{align*} \hat{\boldsymbol{\mu}}_{i} &=\frac{1}{n_{i}} \sum_{l \in \mathcal{T}_{i}} \mathbf{x}_{l}, \quad i \in\{0,1\} \\ \widehat{\mathbf{\Sigma}}_{i} &=\frac{1}{n_{i}-1} \sum_{l \in \mathcal{T}_{i}}\left(\mathbf{x}_{l}-\hat{\boldsymbol{\mu}}_{i}\right)\left(\mathbf{x}_{l}-\hat{\boldsymbol{\mu}}_{i}\right)^{T}, \quad i \in\{0,1\}     \end{align*}
In case the number of samples $n_0$ or $n_1$ is less than the number of features, the use of the sample covariance matrix as plug-in estimator is not permitted since the inverse could not be defined. A popular approach to circumvent this issue is to consider a regularized estimator of the inverse of the covariance matrix given by  
\begin{align} \mathbf{H}_i(\gamma) &=\left(\mathbf{I}_{p}+\gamma \widehat{\mathbf{\Sigma}}_{i}\right)^{-1} \quad ,i \in \{0,1\}    \label{4}\end{align}
where $\gamma$  is a  regularization parameter, which serves to shrink the sample covariance matrix towards identity. Replacing $\boldsymbol{\Sigma}_i^{-1}$ by ${\bf H}_i(\gamma)$ yields the following classification rule for the traditional R-QDA:
\begin{align}\nonumber
 &\widehat{W}^{R-Q D A}(\mathbf{x}) \nonumber \\ & =\frac{1}{2} \log \frac{\left|\mathbf{H}_{0}(\gamma)\right|}{\left|\mathbf{H}_{1}(\gamma)\right|}-\frac{1}{2}\left(\mathbf{x}-\hat{\boldsymbol{\mu}}_{0}\right)^{T} \mathbf{H}_{0}(\gamma)\left(\mathbf{x}-\hat{\boldsymbol{\mu}}_{0}\right) \nonumber \\&+\frac{1}{2}\left(\mathbf{x}-\hat{\boldsymbol{\mu}}_{1}\right)^{T} \mathbf{H}_{1}(\gamma)\left(\mathbf{x}-\hat{\boldsymbol{\mu}}_{1}\right)-\log \frac{\pi_{1}}{\pi_{0}} \label{5}
\end{align}
The classifier R-QDA assigns wrongly observation ${\bf x}$ if $\widehat{W}^{R-Q D A}(\mathbf{x})<0$ when ${\bf x}\in\mathcal{C}_0$ or if $\widehat{W}^{R-Q D A}(\mathbf{x})>0$ when ${\bf x}\in\mathcal{C}_1$. 
Conditioning on the training sample $\mathcal{T}_{i}, i \in\{0,1\}$, the classification error associated with class $\mathcal{C}_i$, is thus given by
\begin{align}\label{7}
\epsilon_{i}^{R-Q D A}=\mathbb{P}\left[(-1)^{i} \widehat{W}^{R-Q D A}(\mathbf{x})<0 | \mathbf{x} \in \mathcal{C}_{i}, \mathcal{T}_{0}, \mathcal{T}_{1}\right]\
\end{align}
which gives the following expression for the total misclassification error probability
\begin{equation}\label{8}
\epsilon^{R-Q D A}=\pi_{0} \epsilon_{0}^{R-Q D A}+\pi_{1} \epsilon_{1}^{R-Q D A}.
\end{equation}
\subsection{Identification of the problems of the R-QDA classifier in imbalanced data settings }

In this section, we unveil several issues related to the use of the classification rule~\eqref{eq:rule} of R-QDA in high dimensional settings. First, we shall recall  that for a classification rule to be able to discriminate observations, it is essential that it presents a non-negligible difference in  distributional behavior when the testing observation changes from one class to the other. 
Clearly, if it behaves similarly for all testing observations, it would not be possible for it to distinguish between observations from different classes. 
This change in behavior should  be reflected by a notable difference in the  expected values of the classification rule when the testing observations belong to class $\mathcal{C}_0$ or $\mathcal{C}_1$, which by reference to \eqref{eq:rule} need to be preferably of opposite signs.   
Consider the normalized classification rule of the traditional R-QDA $\frac{1}{\sqrt{p}} \widehat{W}^{R-Q D A}(\mathbf{x}))$, and denote by $\overline{S}_i$ and $\overline{V}_i$ its expected value and its variance taken over the distribution of the testing observation ${\bf x}$ when it belongs to class $\mathcal{C}_i$. Letting $\boldsymbol{\mu}=\boldsymbol{\mu}_1-\boldsymbol{\mu}_0$,   $\overline{S}_i$ and $\overline{V}_i$ are thus given by: 
\begin{align}
	\overline{S}_i&=\frac{1}{2\sqrt{p}}\log\frac{|{\bf H}_0(\gamma)|}{|{\bf H}_1(\gamma)|} -\frac{1}{2\sqrt{p}}\left(\boldsymbol{\mu}_i-\hat{\boldsymbol{\mu}}_0\right)^{T}{\bf H}_0(\gamma)\left(\boldsymbol{\mu}_i-\hat{\boldsymbol{\mu}}_0\right)\nonumber \\ &-\frac{1}{\sqrt{p}}\log\frac{\pi_1}{\pi_0} \nonumber +\frac{1}{2\sqrt{p}}\left(\boldsymbol{\mu}_i-\hat{\boldsymbol{\mu}}_1\right)^{T}{\bf H}_1(\gamma)\left(\boldsymbol{\mu}_i-\hat{\boldsymbol{\mu}}_1\right)\\ & -\frac{1}{2\sqrt{p}}{\rm Tr} \big[\boldsymbol{\Sigma}_i{\bf H}_0(\gamma)\big]+\frac{1}{2\sqrt{p}}{\rm Tr} \big[\boldsymbol{\Sigma}_i{\bf H}_1(\gamma)\big]\label{eq:s_i}\\
	\overline{V}_i&=\frac{1}{2p}{\rm Tr}\big(\left({\bf H}_1(\gamma)-{\bf H}_0(\gamma)\right)\boldsymbol{\Sigma}_i({\bf H}_1(\gamma)-{\bf H}_0(\gamma)\boldsymbol{\Sigma}_i\big) \nonumber \\
	&+\left((\boldsymbol{\mu}_i^{T}-\hat{\boldsymbol{\mu}}_1^{T}){\bf H}_1(\gamma)+(-\boldsymbol{\mu}_i^{T}+\hat{\boldsymbol{\mu}}_0^{T}){\bf H}_0(\gamma)\right)\boldsymbol{\Sigma}_i\nonumber \\ &\left({\bf H}_1(\gamma)(\boldsymbol{\mu}_i-\hat{\boldsymbol{\mu}}_1)+{\bf H}_0(\gamma)(-\boldsymbol{\mu}_i+\hat{\boldsymbol{\mu}}_0)\right)
\end{align}
At this point, we shall recall that $\overline{S}_i$ and $\overline{V}_i$ are still random since they depend on the training data which are assumed to be drawn independently from the distribution associated with each class. At first sight, the expressions of $\overline{S}_i$ and $\overline{V}_i$ are complicated and it does not seem that too much information can be drawn from them. To gain insights into their behavior in high dimensional settings, we  consider the regime in which $n_0$, $n_1$  and $p$ are large and commensurable with $n_0$ and $n_1$ not asymptotically comparable, ($\frac{n_0}{n_1}\to \ell \neq 1$) and assume additionally that 
the spectral norms of $\boldsymbol{\Sigma}_i, i=\left\{0,1\right\}$ do not grow with  $p$ while $\|\boldsymbol{\mu}_1-\boldsymbol{\mu}_0\|$ scales at most like $O(p^{\frac{1}{4}})$.  Under these assumptions, it is easy to see that $\overline{S_i}$ and $\overline{V}_i$ satisfy:
\begin{align}
&\hspace{-0.4cm}\overline{S}_i \nonumber\\&\hspace{-0.5cm}=\underbrace{\frac{1}{2\sqrt{p}}\!\!\log\frac{|{\bf H}_0(\gamma)|}{|{\bf H}_1(\gamma)|}\!\!-\frac{1}{2\sqrt{p}}\!\!{\rm Tr} \big[\boldsymbol{\Sigma}_i{\bf H}_0(\gamma)\big]\!\! +\frac{1}{2\sqrt{p}}\!\!{\rm Tr} \big[\boldsymbol{\Sigma}_i{\bf H}_1(\gamma)\big]}_{O_p(\sqrt{p})}\!\!\nonumber \\&+O_p(1) 
	\label{eq:Si}\\
		\overline{V}_i&=O_p(1).
\end{align}
where we recall that $X=O_p(p^\alpha)$ means that $\frac{1}{p^{\alpha}}X$ is bounded in probability (See \cite{vaart98book} for the formal definition of the notation $O_p(.)$).  
Several important remarks are in order regarding \eqref{eq:Si}. First, we note that the  prior probabilities $\pi_1$ and $\pi_0$ do not play asymptotically any role in the classification, since the term $\frac{1}{2\sqrt{p}}\log \frac{\pi_1}{\pi_0}$ tends to zero. Hence, the information regarding the prior probabilities is asymptotically lost in the quantities $\overline{S}_i,i=0,1$.
Second, one can easily see that if the distance between the covariances is such that $\frac{1}{\sqrt{p}}{\rm Tr}\big[\boldsymbol{\Sigma}_1{\bf H}_0\big]-\frac{1}{\sqrt{p}}{\rm Tr}\big[\boldsymbol{\Sigma}_0{\bf H}_0\big]=O_p(1)$ and  $\frac{1}{\sqrt{p}}{\rm Tr}\big[\boldsymbol{\Sigma}_1{\bf H}_1\big]-\frac{1}{\sqrt{p}}{\rm Tr}\big[\boldsymbol{\Sigma}_0{\bf H}_1\big]=O_p(1)$ which occurs for instance when $\boldsymbol{\Sigma}_1-\boldsymbol{\Sigma}_0$ has at most rank $\sqrt{p}$ \cite{kh}, the quantities $\overline{S}_i$ for $i=0,1$ are given by:
\begin{align}
	\overline{S}_i=&\frac{1}{2\sqrt{p}}\log\frac{|{\bf H}_0(\gamma)|}{|{\bf H}_1(\gamma)|}-\frac{1}{2\sqrt{p}}{\rm Tr}\big[\boldsymbol{\Sigma}_1{\bf H}_0(\gamma)\big]\nonumber\\&+\frac{1}{2\sqrt{p}}{\rm Tr} \big[\boldsymbol{\Sigma}_1{\bf H}_1(\gamma)\big]+O_p(1), \ \ i=\{0,1\}. \label{eq:sip}
\end{align}
From \eqref{eq:sip}, it appears that the highest order of $\overline{S}_i$ is  $O_p(\sqrt{p})$ but is non-informative, being the same for all testing observations regardless of the class to which they belong. Moreover as the variance is $O_p(1)$, from the Chebyshev's inequality (applied conditioning on the training samples), one can deduce that  R-QDA would keep the same sign for the majority of the testing observations irrespective of their corresponding classes. 

To visually illustrate this result, we display in Figure \ref{fig:sub-first} and Figure \ref{fig:sub-second} the histograms of the QDA statistic  in \eqref{eq:rule}, and that of R-QDA in \eqref{5} when applied to testing observations from both classes. As can be seen, using true statistics, QDA presents a clear change in distribution  that visually should allow distinction between both classes. However, when using  R-QDA , there is an important overlap between the histograms associated with both classes, with all realizations presenting the same sign. By reference to the decision rule in \eqref{eq:rule}, this should lead to the R-QDA  assigning all observations to the same class.   

The reason why the same behavior is not encountered when the same number of  training samples is used for both classes lies in that under this setting, sample covariance matrices of both classes are computed based on the same  number of training samples. The scores associated with the two classes are thus comparable, and as such their difference which form the R-QDA statistic, cancels out the non-informative estimation induced noise and  keep asymptotically the relevant information to classification.   On the opposite, when both classes do not have the same training samples, the scores associated with each class contain estimation induced noises which are not of the same level. The statistic of R-QDA resulting from computing the difference between these scores will thus be essentially at its highest order a non-informative quantity caused by this difference in estimation quality of the covariance matrices. As shall be shown next, the use of RMT tools theoretically confirms this intuition, and most importantly, allowed us to propose an RMT-improved QDA classifier, outfitted with two regularization parameters as well as a modified bias, that will be carefully chosen so that they minimize the misclassification error rate. More formally, the classification rule associated with the proposed classifier is given by:     



\begin{figure}
\begin{subfigure}{0.5\textwidth}
\includegraphics[width=1\linewidth, height=5cm]{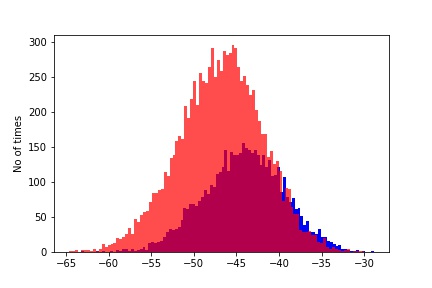} 
  \caption{R-QDA based on regularized covariance estimate}
  \label{fig:sub-first}
\end{subfigure}
\begin{subfigure}{0.5\textwidth}
\includegraphics[width=1\linewidth, height=5cm]{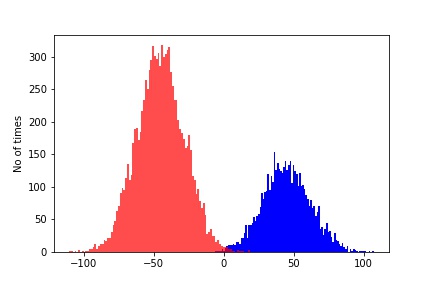}
  \caption{QDA classifier based on true statistics }
  \label{fig:sub-second}
\end{subfigure}

\caption{Histogram of the classification rule for the case with regularized covariance estimate where $\gamma_{0}=10$ and the case with perfect knowledge of the covariance matrices. We consider $p= 1000$ features with unbalanced training size where $n_{0}=500, n_{1}=1000$, $\boldsymbol{\Sigma}_{0} =10 \times \mathbf{I}_{p}$, $\mathbf{\Sigma}_{1}=\mathbf{\Sigma}_{0},\boldsymbol{\mu}_{0}=\mathbf{0}_{p \times 1}$ and  $\boldsymbol{\mu}_{1}=\boldsymbol{\mu}_{0}+\frac{3}{\sqrt{p}} \mathbf{1}_{p \times 1}$. The  testing set is of size 5000 and 10000 samples for the first and second class respectively.}
\label{fig:fig}
\end{figure}
\begin{align}
	& \widehat{W}^{R-Q D A^{\rm imp}}(\mathbf{x})\label{eq:improved} =
\frac{-\theta}{2} \sqrt{p}-\frac{1}{2}\left(\mathbf{x}-\hat{\boldsymbol{\mu}}_{0}\right)^{T} \mathbf{H}_{0}(\gamma_{0})\left(\mathbf{x}-\hat{\boldsymbol{\mu}}_{0}\right)\nonumber \\
  &+\frac{1}{2}\left(\mathbf{x}-\hat{\boldsymbol{\mu}}_{1}\right)^{T} \mathbf{H}_{1}(\gamma_{1})\left(\mathbf{x}-\hat{\boldsymbol{\mu}}_{1}\right)
\end{align}
where 1) $\gamma_{0}$ and $\gamma_{1}$ are two regularization parameters weighting the sample covariance matrix of each class and carefully devised so that the expected value  $\mathbb{E}_{{\bf x}}\left[\frac{1}{\sqrt{p}}\widehat{W}^{R-Q D A^{\rm imp}}(\mathbf{x})\right]$ when ${\bf x}\in\mathcal{C}_0$ or $\mathcal{C}_1$ are $O_p(1)$  and reflects  the class under consideration, and 2) $\theta$  is a bias term that will be set to the value that minimizes the asymptotic classification error rate. 
\section{Design of the improved R-QDA classifier}
\label{section:improved_qda}
In this section, we propose an improved design of the R-QDA classifier that fixes  the aforementioned issues met in imbalanced settings. The design will be based on performing an asymptotic analysis of the statistics in \eqref{eq:improved} under the following asymptotic regime,

\noindent\textbf{Assumption. \hyperlink{1}{1}} (Data scaling). $\frac{p}{n} \rightarrow c \in(0, \infty)$ and $\frac{n_0}{n_1}\rightarrow \ell $ \\
\textbf{Assumption. \hyperlink{2}{2}} (Mean scaling). $\left\|\boldsymbol{\mu}_{0}-\boldsymbol{\mu}_{1}\right\|^{2}=O(\sqrt{p})$\\
\textbf{Assumption. \hyperlink{3}{3}} (Covariance scaling). $\left\|\boldsymbol{\Sigma}_{i}\right\|=\Theta(1)$, $i=0,1$\\
\textbf{Assumption. \hyperlink{4}{4}.} Matrix $\boldsymbol{\Sigma}_{0}-\boldsymbol{\Sigma}_{1}$ has exactly $\Theta(\sqrt{p})$ eigenvalues of order $\Theta(1)$. The remaining eigenvalues are  $O\left(\frac{1}{\sqrt{p}}\right)$.

Assumption 1 and 3 are standard and are often used  to describe a growth regime in which the number of features scales comparably with that of samples and the spectral norm of both covariance matrices remain bounded. Note, however, that Assumption 1 is more general than the one considered in  the work of \cite{kh}, as it accounts for  imbalanced settings in which $\frac{n_0}{n_1}\to \ell\neq 1$. Assumption 2 provides the minimal distance scaling between the mean vectors so that they can be used to discriminate between both classes, \cite{couillet} . Finally Assumption 4, introduced in \cite{kh} specifies the difference between covariances that suffices on its own (regardless of the condition on the mean vectors) to inform  on the class of the testing observation.   

Under the asymptotic regime specified by Assumptions 1-4 and along the same lines as in \cite{kh}, we analyze  the classification error rate of the proposed classifier based on the classification rule \eqref{eq:improved}. Before presenting the corresponding result, we shall first introduce the following notations which defines deterministic objects that naturally appears when using random matrix theory results.

For $i=0,1$, let $\delta_i$ be the unique positive solution to the following equation:
\begin{equation}
\delta_i=\frac{1}{n_i}{\rm Tr} \left[\boldsymbol{\Sigma}_{i}\left(\mathbf{I}_{p}+\frac{\gamma_i}{1+\gamma_i \delta_{i}} \boldsymbol{\Sigma}_{i}\right)^{-1}\right]\label{10}
\end{equation}
The existence and uniqueness of $\delta_i$ follows from standard results in random matrix theory \cite{walid_new}. For $i=0,1$, we also define matrices ${\bf T}_i$, as:
\begin{equation}
\mathbf{T}_{i}=\left(\mathbf{I}_{p}+\frac{\gamma_i}{1+\gamma_i \delta_{i}} \boldsymbol{\Sigma}_{i}\right)^{-1}\label{11}
\end{equation}
and the scalars $\phi_i$ and $\tilde{\phi}_i $ as:
\begin{equation}
	\phi_{i}=\frac{1}{n_{i}} {\rm Tr}\big[\mathbf{\Sigma}_{i}^{2} \mathbf{T}_{i}^{2}\big], \quad \tilde{\phi}_{i}=\frac{1}{\left(1+\gamma_{i} \delta_{i}\right)^{2}}\label{12}
\end{equation}
With these notations at hand, we are now in position to state the first asymptotic result:\\
\noindent 
{\bf  \hypertarget{th:main}{Theorem 1}} {\it
	Under Assumption 1-4, and assuming that the regularization parameters $\gamma_0$ and $\gamma_1$ are $\Theta(1)$, for $i=\{0,1\}$,  the classification error rate associated with class $\mathcal{C}_i$ defined as $
\epsilon_{i}^{\rm imp }=\mathbb{P}\left[(-1)^i\widehat{W}^{R-QDA^{\rm imp}}({\bf x})<0 \ | \ {\bf x}\in\mathcal{C}_i,\mathcal{T}_0,\mathcal{T}_1 \right]
$ 
	satisfies:
\begin{equation}
	\epsilon_{i}^{\rm imp }-\Phi\left((-1)^{i} \frac{\overline{\xi}_{i}-\overline{b_{i}}}{\sqrt{2 \overline{B_{i}}+4\overline{r}_i}}\right) \stackrel{p}{\rightarrow} 0\label{13}
\end{equation}
where 
\begin{align}
	\overline{\xi}_{i} &\triangleq \frac{1}{\sqrt{p}}\left[(-1)^{i+1} \boldsymbol{\mu}^{T} \mathbf{T}_{1-i} \boldsymbol{\mu}\right]+\theta \quad \textnormal{with} \ \ \boldsymbol{\mu}=\boldsymbol{\mu}_1-\boldsymbol{\mu}_0\label{15}\\
	\overline{b}_{i}&=\frac{1}{\sqrt{p}} {\rm Tr} \boldsymbol{\Sigma}_{i}\left(\mathbf{T}_{1}-\mathbf{T}_{0}\right)\\
	\overline{B_{i}}&=\frac{\phi_{i}}{1-\gamma_{i}^{2} \phi_{i} \tilde{\phi}_{i}} \frac{n_{i}}{p}+\frac{1}{p} {\rm Tr}\big[ \boldsymbol{\Sigma}_{i}^{2} \mathbf{T}_{1-i}^{2}\big]\nonumber -\frac{2}{p} {\rm Tr} \big[\boldsymbol{\Sigma}_{i} \mathbf{T}_{1} \boldsymbol{\Sigma}_{i} \mathbf{T}_{0}\big]\\
	&+\frac{n_{i}}{p} \frac{\gamma_{1-i}^{2} \tilde{\phi}_{1-i}}{1-\gamma_{1-i}^{2} \phi_{1-i} \tilde{\phi}_{1-i}}\left(\frac{1}{n_{i}} {\rm Tr}\big[ \boldsymbol{\Sigma}_{i} \boldsymbol{\Sigma}_{1-1} \mathbf{T}_{1-i}^{2}\big]\right)^{2} \\  
		\overline{r}_i&=\frac{\frac{1}{p}\boldsymbol{\mu}^{T}\boldsymbol{\Sigma}_{1-i}{\bf T}_{1-i}^2\boldsymbol{\mu}}{1-\gamma_{1-i}^2\phi_{1-i}\tilde{\phi}_{1-i}}
\end{align}
}
\noindent

{\bf Proof}.
	We will provide only a sketch of proof since it follows along the same lines as in \cite{kh}. To begin with, note that $\frac{1}{\sqrt{p}} \widehat{W}^{R-QDA^{\rm imp}}({\bf x})$ is a quadratic form on the testing observation ${\bf x}$ which, when it belongs to class $\mathcal{C}_i, i=\{0,1\}$ is assumed to follow Gaussian distribution with mean $\boldsymbol{\mu}_i$ and covariance $\boldsymbol{\Sigma}_i$. By Lyapunov's central limit theorem (\cite{billingsley}), when ${\bf x}$ is in $\mathcal{C}_i$, for $i\in\{0,1\}$, $\frac{1}{\sqrt{p}} \widehat{W}^{R-QDA^{\rm imp}}({\bf x})$ satisfies: 
	\begin{equation}
	\frac{1}{\sqrt{\tilde{V_i}}}\left(\frac{2}{\sqrt{p}}\widehat{W}^{R-QDA^{\rm imp}}({\bf x})-\tilde{S}_i\right) \overset{d}{\to}\mathcal{N}(0,1)
		\label{eq:conv_distribution}
	\end{equation}
	where $\tilde{S}_i$ and $\tilde{V}_i$ are given by:
	\begin{align}
		\tilde{S}_i&=-\theta-\frac{1}{\sqrt{p}}{\rm Tr}\boldsymbol{\Sigma}_i{\bf H}_0(\gamma_0)+\frac{1}{\sqrt{p}}{\rm Tr}\boldsymbol{\Sigma}_i{\bf H}_1(\gamma_1)\nonumber\\
		&-\frac{1}{\sqrt{p}}(\boldsymbol{\mu}_i-\hat{\boldsymbol{\mu}}_0)^{T}{\bf H}_0(\gamma_0)(\boldsymbol{\mu}_i-\hat{\boldsymbol{\mu}}_0)\nonumber \\
		&+\frac{1}{\sqrt{p}}(\boldsymbol{\mu}_i-\hat{\boldsymbol{\mu}}_1)^{T}{\bf H}_1(\gamma_1)(\boldsymbol{\mu}_i-\hat{\boldsymbol{\mu}}_1)\\
		\tilde{V}_i&=\frac{2}{p}{\rm Tr}\left({\bf H}_1(\gamma_1)-{\bf H}_0(\gamma_0)\right)\boldsymbol{\Sigma}_i\left({\bf H}_1(\gamma_1)-{\bf H}_0(\gamma_0)\right)\nonumber\\
			&+\frac{4}{p}\left((\boldsymbol{\mu}_i^{T}-\hat{\boldsymbol{\mu}}_1^{T}){\bf H}_1(\gamma_1)+(-\boldsymbol{\mu}_i^{T}+\hat{\boldsymbol{\mu}}_0^{T}){\bf H}_0(\gamma_0)\right)\boldsymbol{\Sigma}_i \nonumber \\ 
			&\left({\bf H}_1(\gamma_1)(\boldsymbol{\mu}_i-\hat{\boldsymbol{\mu}}_1)+{\bf H}_0(\gamma_0)(-\boldsymbol{\mu}_i+\hat{\boldsymbol{\mu}}_0)\right)
	\end{align}
	From \eqref{eq:conv_distribution}, and using Lemma 2.11 in \cite{vanderVaart98} for $i=\{0,1\}$, we can easily see that:
	\begin{equation}
	\epsilon_i^{\rm imp}-\Phi\left((-1)^i\left(-\frac{\tilde{S}_i}{\sqrt{\tilde{V}_i}}\right)\right)\to 0. 
		\label{eq:eps}
	\end{equation}
	It follows using standard results from random matrix theory in \cite{hachem2013} that:
\begin{equation}
	\left(\frac{1}{\sqrt{p}}{\rm Tr}\boldsymbol{\Sigma}_i{\bf H}_1(\gamma_1)-\frac{1}{\sqrt{p}}{\rm Tr}\boldsymbol{\Sigma}_i{\bf H}_0(\gamma_0)\right)- \overline{b}_i\overset{a.s}{\to} 0 \label{eq:bi}
\end{equation}
	Moreover, 
	\begin{equation}
	\frac{1}{\sqrt{p}} \left(\boldsymbol{\mu}_i-\hat{\boldsymbol{\mu}}_i\right)^{T}{\bf H}_i(\gamma_i) \left(\boldsymbol{\mu}_i-\hat{\boldsymbol{\mu}}_i\right)\overset{a.s.}{\to} 0
		\label{eq:mu}
	\end{equation}
	while,
	\begin{align}
	&\frac{1}{\sqrt{p}} \left(\boldsymbol{\mu}_i-\hat{\boldsymbol{\mu}}_{1-i}\right)^{T}{\bf H}_{1-i}(\gamma_{1-i}) \left(\boldsymbol{\mu}_i-\hat{\boldsymbol{\mu}}_{1-i}\right) \nonumber \\ & - \frac{1}{\sqrt{p}}\boldsymbol{\mu}^{T}{\bf T}_{1-i}\boldsymbol{\mu} \overset{a.s.}{\to} 0 
	\label{eq:si_b}
	\end{align}
	Putting \eqref{eq:bi}, \eqref{eq:mu} and \eqref{eq:si_b} together, we obtain:
	\begin{equation}
	-\tilde{S}_i-\left(-\overline{\xi}_i+\overline{b}_i\right)\overset{a.s.}{\to} 0. 
		\label{eq:mean}
	\end{equation}
	The variance quantity can be treated similarly, and we can prove based on the results in \cite{kh} that:
	$$
	\frac{1}{p}{\rm Tr}({\bf H}_1(\gamma_1)-{\bf H}_0(\gamma_0))\boldsymbol{\Sigma}_i({\bf H}_1(\gamma_1)-{\bf H}_0(\gamma_0))-\overline{B}_i\overset{a.s.}{\to} 0
	$$
	and 
\begin{align}
	&\left((\boldsymbol{\mu}_i^{T}-\hat{\boldsymbol{\mu}}_1^{T}){\bf H}_1(\gamma_1)+(-\boldsymbol{\mu}_i^{T}+\hat{\boldsymbol{\mu}}_0^{T}){\bf H}_0(\gamma_0)\right)\boldsymbol{\Sigma}_i  \nonumber \\ 
	& \left({\bf H}_1(\gamma_1)(\boldsymbol{\mu}_i-\hat{\boldsymbol{\mu}}_1)+{\bf H}_0(\gamma_0)(-\boldsymbol{\mu}_i+\hat{\boldsymbol{\mu}}_0)\right)-p\overline{r}_i\overset{a.s.}{\to} 0. 
\end{align}
	Hence,
	\begin{equation}
	\tilde{V}_i-\left(2\overline{B}_i+4\overline{r}_i\right)\overset{a.s.}{\to} 0. 
		\label{eq:variance}
	\end{equation}
	Replacing $-\tilde{S}_i$ and $\tilde{V}_i$ by their deterministic equivalents in \eqref{eq:mean} and \eqref{eq:variance} in \eqref{eq:eps}, we obtain the desired convergence in \eqref{13}.   

\noindent\textbf{Remark:}
 Under Assumption \hypertarget{4}{4}, it can be shown that $\overline{B_{i}}$ can asymptotically be simplified to
  $${\overline{B_{i}} \triangleq \frac{2n_{i}}{p}
	\frac{\gamma^{2}_{i}\tilde{\phi}_{i}\phi_{i}^{2}}{1-\gamma^{2}_{i} \phi_{i} \tilde{\phi}_{i}}}+\Theta(\frac{1}{\sqrt{p}})$$
	and that $\overline{B}_1=\overline{B}_0+\Theta(\frac{1}{\sqrt{p}})$. 
Moreover, the term $\overline{r}_i$ is $O(\frac{1}{\sqrt{p}})$ and as such converges to zero as $p,n$ grow to infinity. However,
in our simulations, we chose to work with the non-simplified expressions for $\overline{B}_i$ and to keep the term $\overline{r}_i$, since 
we observed that in doing so a better accuracy is obtained in finite-dimensional  simulations. 

The result of \hyperlink{th:main}{Theorem 1} allows to provide guidelines on how to choose $\gamma_0$ and $\gamma_1$ and the optimal bias $\theta$. As discussed before, the design should require the mean of the classification rule to be $\Theta(1)$ and to reflect the class under consideration. This mean is represented in the asymptotic expression of the classification error rate by the quantity $\overline{\xi}_i-\overline{b}_i$ which is $\Theta(\sqrt{p})$ for arbitrary $\gamma_0$ and $\gamma_1$  as $\overline{b}_i=\Theta(\sqrt{p})$ and $\overline{\xi}_i=\Theta(1)$. Moreover, the class of the testing observation is not reflected in $\overline{b}_i$ since under Assumption 3-4, in case $\overline{b}_i=\Theta(\sqrt{p})$, $\overline{b}_i=\frac{1}{\sqrt{p}}{\rm Tr} \boldsymbol{\Sigma}_1({\bf T}_1-{\bf T}_0)+O(1)$ and as such up to a quantity of order $O(1)$, $\overline{b}_0$ and $\overline{b}_1$ are equal.  To solve this issue, we need to design $\gamma_1$ and $\gamma_0$ such that for $i=\{0,1\}$, $\overline{b}_i$ is $\Theta(1)$ or equivalently,
\begin{equation}
	\frac{1}{p}{\rm Tr} \big[\boldsymbol{\Sigma}_1({\bf T}_1-{\bf T}_0)\big]=\Theta(\frac{1}{\sqrt{p}})
	\label{eq:necessary}
\end{equation}
so that $\overline{b}_0$ becomes different from $\overline{b}_1$ at its highest order.  To this end, we prove that it suffices to select the regularization parameter associated with the class with the largest number of samples as: \\
\noindent
{\bf  \hypertarget{th:reg}{Theorem 2}} {\it	Under assumption 1-4, and assume that $n_1>n_0$,  if \begin{equation}\gamma_{1}=\frac{\gamma_{0}}{1-\left(\frac{1}{n}_{1}-\frac{1}{n_{0}}\right) \gamma_{0} {\rm Tr} \big[ \boldsymbol{\Sigma}_{0} \mathbf{T}_{0}\big]},\label{eq:gamma}\end{equation} where $\gamma_0$ is fixed to a given constant
	then $\overline{b}_i=\Theta(1)$. 
}

\noindent

{\bf Proof}.	See Appendix A.\\
%
%
%
%
It is worth mentioning that in the balanced case, plugging $n_0=n_1$ into \eqref{eq:gamma} yields $\gamma_1=\gamma_0$. 
This shows that When the same number of training samples is used across classes, it is not necessary to regularize both sample covariance matrices with different regularization parameters.  

Now with this choice of the regularization parameters being set, it remains to select the optimal bias $\theta$. This can be chosen so that the asymptotic classification error rate given by: 
$$
\overline{\epsilon}=\pi_0\Phi\left(-\frac{\overline{\xi}_0-\overline{b}_0}{\sqrt{2\overline{B}_0}}\right)+\pi_1\Phi\left(-\frac{\overline{\xi}_1-\overline{b}_1}{\sqrt{2\overline{B}_1}}\right)
$$
is minimized.  \\

\noindent
{\bf  \hypertarget{th:theta}{Theorem 3}} {\it
The optimal bias that allows to minimize the asymptotic classification error rate is given by:
\begin{align}
    \theta^{*}=\frac{\beta_{1}-\beta_{0}}{2}-\frac{2\alpha^{2}}{\beta_{1}+\beta_{0}}\log(\frac{\pi_{1}}{\pi_{0}})
	\label{eq:theta}
\end{align}
where
\begin{align*}
        \begin{cases}\beta_{0}=\frac{1}{\sqrt{p}}\left[- \boldsymbol{\mu}^{T} \mathbf{T}_{1} \boldsymbol{\mu}\right]-\frac{1}{\sqrt{p}} {\rm Tr}\big[ \boldsymbol{\Sigma}_{0}\left(\mathbf{T}_{1}-\mathbf{T}_{0}\right)\big]\\
    \beta_{1}= \frac{1}{\sqrt{p}}\left[- \boldsymbol{\mu}^{T} \mathbf{T}_{0} \boldsymbol{\mu}\right]+\frac{1}{\sqrt{p}} {\rm Tr}\big[ \boldsymbol{\Sigma}_{1}\left(\mathbf{T}_{1}-\mathbf{T}_{0}\right)\big]\\
    \alpha=\sqrt{2 \overline{B_{0}}}
    \end{cases}
\end{align*}
}
\noindent

{\bf Proof}. See Appendix B.\\
Before proceeding further, it is important to note that thanks to the careful choice of the regularization parameters $\gamma_0$ and $\gamma_1$ provided in \hyperlink{th:reg}{Theorem 2}, the term   $\frac{1}{\sqrt{p}} {\rm Tr}\big[ \boldsymbol{\Sigma}_{i}\left(\mathbf{T}_{1}-\mathbf{T}_{0}\right)\big]$ is $\Theta(1)$ for i $\in \{0,1\}$, 
Additionally, it can be shown easily that the term $\frac{1}{\sqrt{p}}\left[- \boldsymbol{\mu}^{T} \mathbf{T}_{i} \boldsymbol{\mu}\right]$ is of order $\Theta(1)$. As a result, both $\beta_{0}$ and $\beta_{1}$ are $\Theta(1)$. \\
On another note, it is worth mentioning that even in the case of balanced classes $n_0=n_1$,  characterized by $\gamma_1=\gamma_0$ as proved in \hyperlink{th:reg}{Theorem 2} , the optimal bias is different from the one traditionally used in R-QDA. As such, the proposed design improves on the traditional R-QDA studied in \cite{kh} even in the balanced case as it optimally adapts the bias term to the situation in which the covariance matrices are not known. \\ 
\hyperlink{th:reg}{Theorem 2} and \hyperlink{th:theta}{Theorem 3} can be used to obtain an optimized design of the proposed R-QDA classifier.
As can be seen, the improved classifier employs only one regularization parameter associated with the class that presents the smallest number of training samples. Assume $\mathcal{C}_0$ is such a class. The regularization parameter associated with the other class  cannot be arbitrarily chosen and should be set as \eqref{eq:gamma}, while the bias is selected according to \eqref{eq:theta}. However, pursuing this design is not possible in practice due to the dependence of \eqref{eq:gamma} and \eqref{eq:theta} on the true covariance matrices. To solve this issue, we propose in the following theorem consistent estimators for the quantities arising in \eqref{eq:gamma} and \eqref{eq:theta} that depend only on the training samples. 

\noindent
{\bf  \hypertarget{th:4}{Theorem 4}} {\it
	Assume $n_1>n_0$ and let $\gamma_0$ be the regularization parameter associated with class $\mathcal{C}_0$. Let $\hat{\delta}_0$ be given by:
	$$
	\hat{\delta}_0=\frac{1}{\gamma_0} \frac{\frac{p}{n_0}-\frac{1}{n_0}{\rm Tr}\big[ {\bf H}_0(\gamma_0)\big]}{1-\frac{p}{n_0}+\frac{1}{n_0}{\rm Tr}\big[{\bf H}_0(\gamma_0)\big]}
	$$
	and define $\hat{\gamma}_1$ as:
	\begin{equation}
	\hat{\gamma}_1=\frac{\gamma_0}{1-\gamma_0\left(\frac{n_0}{n_1}\hat{\delta}_0-\hat{\delta}_0\right)}
		\label{eq:hatgamma}
	\end{equation}
	Then,
	$$
	\hat{\gamma}_1-\gamma_1\stackrel{as}{\rightarrow} 0
	$$
	where $\gamma_1$ is given in \eqref{eq:gamma}.
	Define $\hat{\beta}_0$, $\hat{\beta}_1$ and $\hat{\alpha}$ as:
	\begin{align}
	\nonumber	 \hat{\beta_{0}} & =-\frac{1}{\sqrt{p}}\left(\hat{\boldsymbol{\mu}}_{0}-\hat{\boldsymbol{\mu}}_{1}\right)^{T} \mathbf{H}_{1}(\hat{\gamma}_1)\left(\hat{\boldsymbol{\mu}}_{0}-\hat{\boldsymbol{\mu}}_{1}\right) \nonumber \\
			&-\frac{1}{\sqrt{p}} {\rm Tr} \left[ \boldsymbol{\hat{\Sigma}}_{0}\mathbf{H}_{1}(\hat{\gamma}_1)\right]+\frac{n_{0}}{\sqrt{p}} \hat{\delta}_{0}\\
			\hat{\beta_{1}} & = -\frac{1}{\sqrt{p}}\left(\hat{\boldsymbol{\mu}}_{0}-\hat{\boldsymbol{\mu}}_{1}\right)^{T} \mathbf{H}_{0}(\gamma_0)\left(\hat{\boldsymbol{\mu}}_{0}-\hat{\boldsymbol{\mu}}_{1}\right)\nonumber \\ 
			&-\frac{1}{\sqrt{p}} {\rm Tr} \left[ \boldsymbol{\hat{\Sigma}}_{1}\mathbf{H}_{0}({\gamma}_0)\right]+\frac{n_{1}}{\sqrt{p}}\hat{\delta}_{1}\\ \nonumber
    \hat{\alpha} & =\sqrt{2 \hat{B_{0}}}
\end{align}
		where $\hat{B}_0$ writes as:
	\begin{align}
		\hat{B}_0&=\left(1+\gamma_{0} \widehat{\delta}_{0}\right)^{4} \frac{1}{p} {\rm Tr} \left[\widehat{\mathbf{\Sigma}}_{0}\nonumber \mathbf{H}_{0}(\gamma_0) \widehat{\mathbf{\Sigma}}_{0} \mathbf{H}_0(\gamma_0)\right]\\
		&-\frac{n_{0}}{p} \widehat{\delta}_{0}^{2}\left(1+\gamma_{0} \widehat{\delta}_{0}\right)^{2}+\frac{1}{p} {\rm Tr}\left[ \widehat{\mathbf{\Sigma}}_{0} \mathbf{H}_{1}(\hat{\gamma}_1) \widehat{\mathbf{\Sigma}}_{0} \mathbf{H}_{1}(\hat{\gamma_1})\right]\nonumber\\
		&-\frac{n_{0}}{p}\left(\frac{1}{n_{0}} {\rm Tr}\left[ \widehat{\mathbf{\Sigma}}_{0} \mathbf{H}_{1}(\hat{\gamma}_1)\right]\right)^{2}\nonumber\\
		&-2\left(1+\gamma_{0} \hat{\delta}_{0}\right)^{2} \frac{1}{p} {\rm Tr}\left[ \widehat{\mathbf{\Sigma}}_{0} \mathbf{H}_{0}(\gamma_0) \widehat{\mathbf{\Sigma}}_{0} \mathbf{H}_{1}(\hat{\gamma}_1)\right]\nonumber
		\\&+\widehat{\delta}_{0}\left(1+\gamma_{0} \widehat{\delta}_{0}\right) \frac{2}{p} {\rm Tr} \left[\widehat{\mathbf{\Sigma}}_{0} \mathbf{H}_{1}(\hat{\gamma_1})\right]\label{eq:B0}
\end{align}
	Let $\hat{\theta}^\star$ be given by:
	\begin{equation}\hat{\theta^{\star}}=\frac{\hat{\beta}_{1}-\hat{\beta}_{0}}{2}-\frac{2\hat{\alpha}^{2}}{\hat{\beta}_{1}+\hat{\beta}_{0}}\log(\frac{\pi_{1}}{\pi_{0}})\label{eq:hattheta}\end{equation}
Then, $$
\widehat{\theta^{\star}}-\theta^\star \stackrel{as}{\rightarrow}0
$$
	where $\theta^\star$ is given in \eqref{eq:theta}.
}
\noindent

{\bf Proof}.
See Appendix C.\\
It is worth mentioning that unlike $\gamma_0$, $\hat{\gamma}_1$ is random. It does not satisfy with equality \eqref{eq:gamma} but 
ensures \eqref{eq:necessary} almost surely. Its use as a replacement of $\gamma_1$ would lead asymptotically to the same results as the improved classifier using $\gamma_1$.   

For the reader convenience, we provide hereafter the algorithm describing the proposed improved QDA classifier:
\vspace{2mm}
\hrule
\vspace{1mm}
\textbf{ Algorithm \hyperlink{alg1}{1}: }Improved design of the R-QDA classifier.  \vspace{1mm}
\hrule
\vspace{1mm}
\textbf{Input} : Assuming $n_1\geq n_0$, let $\gamma_0$ the regularization parameter associated with class $\mathcal{C}_0$, $\mathcal{T}_0=\left\{{\bf x}_l\right\}_{l=1}^{n_0}$ training samples in $\mathcal{C}_0$ and $\mathcal{T}_1=\left\{{\bf x}_l\right\}_{l=n_0+1}^{n=n_0+n_1}$\\
\textbf{output} : Estimation of the parameters $\gamma_1$ and $\theta^\star$ to be plugged in \eqref{eq:improved}
  \begin{enumerate}
 \item Compute $\hat{\gamma}_1$ as in \eqref{eq:hatgamma}\;
 \item Compute $\hat{\theta}$ as in \eqref{eq:hattheta}\;
\item 	Return $\hat{\theta}$ and $\hat{\gamma}_1$ that will be plugged in the classification rule \eqref{eq:improved}
\end{enumerate}
\hrule
\vspace{0.1in}
The improved design described in Algorithm \hypertarget{alg1}{1} depends on the regularization parameter $\gamma_0$ associated with the class with the smallest number of training samples. One possible way to adjust this parameter is to resort to a traditional cross-validation approach which consists in estimating, based on a  set of testing data, the classification error rate for each candidate of the regularization parameter $\gamma_0$. Such an approach presents several drawbacks. First, it is computationally expensive and  is way sub-optimal  as it could only  test few values of $\gamma_0$. As an alternative we propose rather to build a consistent estimator of the classification error rate based on results from random matrix theory, which can later assist in the setting of the regularization parameter $\gamma_0$. This is the objective of the following theorem:   

%

\noindent
{\bf  \hypertarget{th:consistent}{Theorem 5}} {\it
	Under Assumptions 1-4, a consistent estimator of the misclassification error rate associated with class $\mathcal{C}_i$ is given by:
	$$
	\hat{\epsilon}_i=\Phi\left((-1)^{i} \frac{\hat{\xi}_{i}-\hat{b}_{i}}{\sqrt{2 \hat{B}_{i}+4\hat{r}_i}}\right)
	$$
	where $\hat{B}_0$ is given in \eqref{eq:B0}, $\gamma_1$ is set to $\hat{\gamma}_1$ and 
\begin{align*}
	\hat{\xi}_i&= \hat{\theta}^{\star}+ (-1)^{i+1}\frac{1}{\sqrt{p}}\left(\hat{\boldsymbol{\mu}}_{0}-\hat{\boldsymbol{\mu}}_{1}\right)^{T} \mathbf{H}_{1-i}(\gamma_{1-i})\left(\hat{\boldsymbol{\mu}}_{0}-\hat{\boldsymbol{\mu}}_{1}\right)\\
	\hat{\delta}_{i}&=\frac{1}{\gamma_{i}} \frac{\big{[} \frac{p}{n_{i}}- \frac{1}{n_{i}} {\rm Tr} \left(\mathbf{H}_{i}(\gamma_i)\right)\big{]}}{1-\frac{p}{n_{i}}+  \frac{1}{n_{i}}{\rm Tr}\left[\mathbf{H}_{i}(\gamma_i)\right] }, \quad i \in \{0,1\}
\\
	\hat{b}_{i}&=\frac{(-1)^{i}}{\sqrt{p}} {\rm Tr}\left[ \hat{\boldsymbol{\Sigma}}_{i} \mathbf{H}_{1-i}({\gamma}_{1-i})\right]+ \frac{(-1)^{i+1}n_{i}}{\sqrt{p}} \hat{\delta}_{i},\hspace{-0.3cm} \quad i \in \{0,1\}\\
	\hat{B}_{1}&=    \left(1+\gamma_{1} \hat{\delta}_{1}\right)^{4} \frac{1}{p} {\rm Tr}\left[ \hat{\mathbf{\Sigma}}_{1} \mathbf{H}_{1}(\hat{\gamma}_1) \hat{\mathbf{\Sigma}}_{1} \mathbf{H}_{1}(\hat{\gamma}_1)\right]\\
	&-\frac{n_{1}}{p} \hat{\delta}_{1}^{2}\left(1+\gamma_{1} \hat{\delta}_{1}\right)^{2}+\frac{1}{p} {\rm Tr}\left[\hat{\mathbf{\Sigma}}_{1} \mathbf{H}_{0}(\gamma_0) \hat{\mathbf{\Sigma}}_{1} \mathbf{H}_{0}(\gamma_0)\right]\\
	&-\frac{n_{1}}{p}\left(\frac{1}{n_1} {\rm Tr}\left[ \hat{\mathbf{\Sigma}}_{1} \mathbf{H}_{0}(\gamma_0)\right]\right)^{2}\\
	&-2\left(1+\gamma_{1} \hat{\delta}_{1}\right)^{2} \frac{1}{p} {\rm Tr}\left[\hat{\mathbf{\Sigma}}_{1} \mathbf{H}_{1} \hat{\mathbf{\Sigma}}_{1} \mathbf{H}_{0}(\gamma_0)\right]\\&+\hat{\delta}_{1}\left(1+\gamma_{1} \hat{\delta}_{1}\right) \frac{2}{p} {\rm Tr}\left[ \hat{\mathbf{\Sigma}}_{1} \mathbf{H}_{0}(\gamma_0)\right]\\
	\hat{r}_i&=\frac{1}{p}(\hat{\boldsymbol{\mu}}_0-\hat{\boldsymbol{\mu}}_1)^{T}{\bf H}_{1-i}(\hat{\gamma}_{1-i})\hat{\boldsymbol{\Sigma}}_i{\bf H}_{1-i}(\hat{\gamma}_{1-i})(\hat{\boldsymbol{\mu}}_0-\hat{\boldsymbol{\mu}}_1)\\
\end{align*}
in the sense that:
	$$
\widehat{\epsilon}_{i}-\epsilon_{i}^{R-Q D A} \stackrel{as}{\rightarrow} 0
$$
}
\noindent

{\bf Proof}.	The proof is based on employing the consistent estimators provided in \cite{kh} and is such omitted. \\



\section{Numerical results}
\label{section:results}
\subsection{Validation with synthetic data}
In this section, we assess the performance of our improved R-QDA classifier and compare it  the with standard QDA classifier in the case of imbalanced training data. To this end, we start by generating synthetic data for both classes that are compliant with the different assumptions used thoughout this work for the sake of validating our theoretical results.  

\begin{figure}[h]
  \centering
    \includegraphics[width=0.78\linewidth]{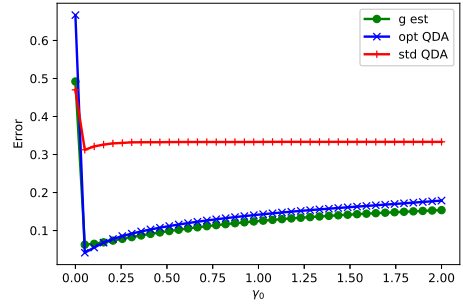}
  \caption{Average misclassification error rate versus the regularization parameter $\gamma_{0}$ using the G-estimator. We consider $p= 1000$ features with unbalanced training size where $n_{0}=2n_{1}$, $\left[\boldsymbol{\Sigma}_{0}\right]=4\mathbf{I}_{p}$,$\mathbf{\Sigma}_{1}=\mathbf{\Sigma}_{0}+3 \mathbf{Q}_{p}\mathbf{D}_{p}\mathbf{Q}_{p}^{T}$, $\mathbf{Q}_{p}\in \mathcal{O}_{n}(R)$,$\mathbf{D}_{p}=diag \left[\mathbf{1}_{\sqrt p}, \mathbf{0}_{(p-\sqrt p)}\right]$ and $\boldsymbol{\mu}_{1}=\boldsymbol{\mu}_{0}+\frac{3}{\sqrt{p}} \mathbf{1}_{p \times 1}$. }
  \label{fig:1}
\end{figure}
\begin{figure}[h]
  \centering
    \includegraphics[width=0.8\linewidth]{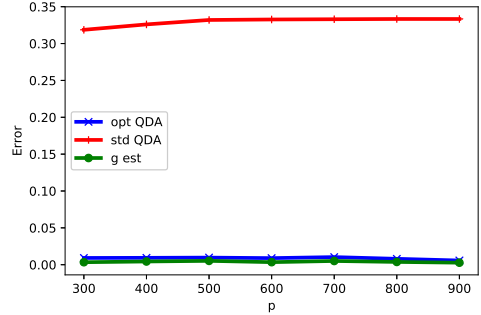}
  \caption{Average misclassification error rate versus the dimension p. We consider $\gamma=1$ with unbalanced training size where $n_{0}=2n_{1}$, $\left[\boldsymbol{\Sigma}_{0}\right]=4\mathbf{I}_{p}$,$\mathbf{\Sigma}_{1}=\mathbf{\Sigma}_{0}+3 \mathbf{Q}_{p}\mathbf{D}_{p}\mathbf{Q}_{p}^{T}$, $\mathbf{Q}_{p}\in \mathcal{O}_{n}(R)$,$\mathbf{D}_{p}=diag \left[\mathbf{1}_{\sqrt p}, \mathbf{0}_{(p-\sqrt p)}\right]$ and $\boldsymbol{\mu}_{1}=\boldsymbol{\mu}_{0}+\frac{3}{\sqrt{p}} \mathbf{1}_{p \times 1}$. }
  \label{fig:2}
\end{figure}
In Figure \ref{fig:1} and  Figure \ref{fig:2}, we plot   the classification error rate of the improved classifier and the traditional R-QDA classifier with respect to the regularization parameter $\gamma_0$ and the features' dimension $p$, respectively. 
As can be seen, we note that the standard R-QDA has a classification error rate that converges to the prior of the most dominant class, which reveals that as expected, it tends to assign all observations to the same class, which in this case coincides with the class that presents the highest number of training samples. On the opposite, the proposed R-QDA classifier presents a much higher performance, making it more suitable to cope with imbalanced settings. We finally note that the consistent estimator based on the results of \hypertarget{th:consistent}{Theorem 5} is accurate and as such can be used to properly adjust the regularization parameter $\gamma_0$.  
\vspace{-0.2cm}
\subsection{Experiment with real data}
\begin{figure}[h]
  \centering
    \includegraphics[width=0.728\linewidth]{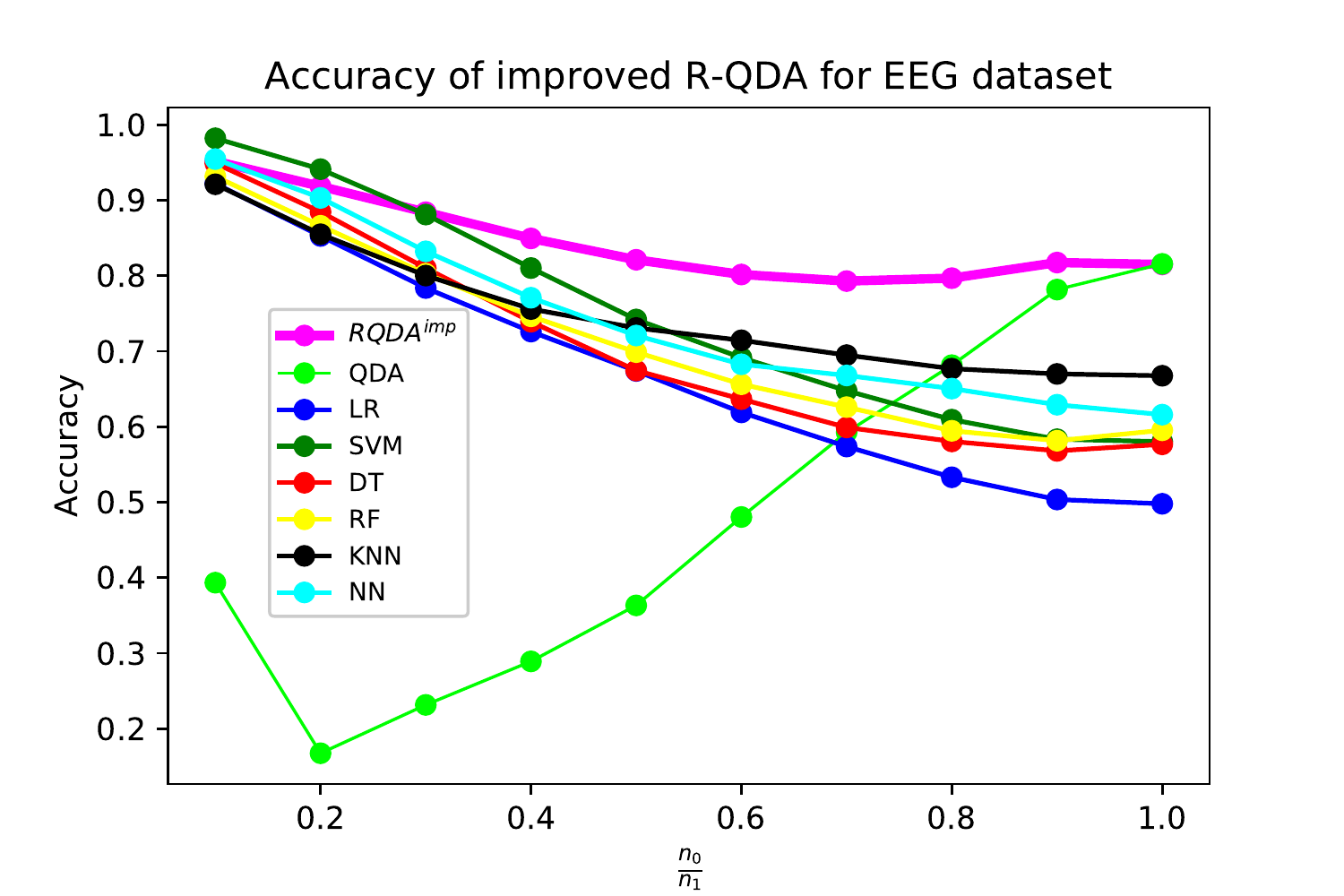}
  \caption{Comparaison between the performance of the our improved RQDA classifier with respect to other machine learning algorithms on the EEG dataset. }
  \label{fig:4}
\end{figure}

\begin{figure}[h]
  \centering
    \includegraphics[width=0.728\linewidth]{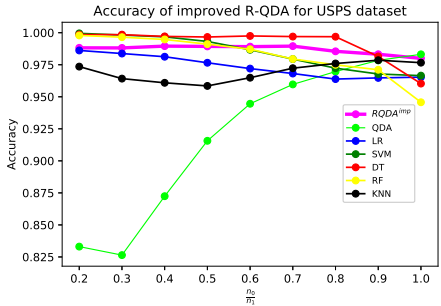}
  \caption{Comparaison between the performance of the our improved RQDA classifier with respect to other machine learning algorithms on the USPS dataset.}
  \label{fig:5}
\end{figure}

In this section, we test the performance of the proposed R-QDA classifier on   the public USPS dataset of handwritten digits\cite{28} and the EEG dataset. The USPS dataset is composed of $42 000$ labeled digit images, and each image has $p= 784$ features represented by $28 \times 28$ pixels. 
The EEG dataset is composed of 5 classes that contain 11,500 observations, and each observation has $p= 178$ features.  We consider the classification of two classes from each dataset composed of $n_0$ and $n_1$ samples.  Based on the results of  \hyperlink{th:consistent}{Theorem 5}, we tune the regularization factor $\gamma_0$ to the value that minimizes the consistent estimate of the  misclassification error rate. The values of $\theta$ and $\hat{\gamma}_1$ are then computed based on \eqref{eq:hatgamma} and \eqref{eq:hattheta}.  
Figure \ref{fig:4} and Figure \ref{fig:5} compare the performance of the proposed  classifier with other state-of-the-art classification algorithms using cross-validation for different ratios of $\frac{n_0}{n_1}$. As seen, our classifier,  termed in the figure ${\textnormal{RQDA}^{\text{imp}}}$, not only outperforms the standard QDA but also other existing classification algorithms. Moreover, it is worth mentioning that the standard R-QDA is the classifier that presents the lowest performance in imbalanced settings corresponding to $\frac{n_0}{n_1}<1$.  This suggests that the use of different regularization across classes in the QDA classification rule along with an adequate tune of the bias makes the R-QDA classifier more robust to the estimation noise of the covariance  matrices in imbalanced settings.   

\section{Conclusion}
\label{section:conclusion}
The traditional view holds that the use of R-QDA leads in general to lower classification performances  than many other existing classification methods, even though  derived from the maximum likelihood principle under a general Gaussian mixture model. 
In this work, we establish that this loss in performance can be attributed to an induced estimation noise of high order that hide the useful information for classification, leading the R-QDA score to behave similarly for all testing observations.  
Based on this analysis,  we propose to modify the design of R-QDA so that it can discriminate efficiently between classes. Our amendment of the R-QDA classifier is based on using two regularization parameters for each class as well as a carefully designed bias that minimizes an asymptotic approximation of the classification performance. We confirm the efficacy of the proposed classifier through the use of a set of numerical results which shows that not only our proposed classifier outperforms the standard R-QDA but also other state-of-the art existing algorithms. Going further, we believe that this work shows that contrary to  common belief, there is still room for improvement of very basic classification methods through a careful study of their behavior using advanced statistical tools.


%

\appendices
\section*{Appendix A.}
\hypertarget{Appendix A Proof Theorem 2}{}
As discussed in the paper, the design of the regularization parameters $\gamma_0$ and $\gamma_1$ should ensure that:
\begin{align}
    \frac{1}{\sqrt{p}} {\rm Tr} \big[\boldsymbol{\Sigma}_{i}\left(\mathbf{T}_{1}-\mathbf{T}_{0}\right)\big]=\Theta(1)
	\label{eq:nece}
\end{align}
where ${\bf T}_i=\left(\mathbf{I}+\gamma_{i} \widetilde{\delta}_{i} \mathbf{\Sigma}_{i}\right)^{-1},$
\text{ with } $\widetilde{\delta}_{i}=\frac{1}{1+\gamma_{i} \delta_{i}}. $ Using the relation ${\bf A}^{-1}-{\bf B}^{-1}={\bf A}^{-1}({\bf B}-{\bf A}){\bf B}^{-1}$ for any two square matrices ${\bf A}$ and ${\bf B}$, \eqref{eq:nece} boils down to:
$$
\frac{1}{\sqrt{p}} {\rm Tr}\left[ \mathbf{\Sigma}_{i} \mathbf{T}_{1}\left(\gamma_{0} \tilde{\delta}_{0} \mathbf{\Sigma}_{0}-\gamma_{1} \tilde{\delta}_{1} \mathbf{\Sigma}_{1}\right) \mathbf{T}_{0}\right]=\Theta(1)
$$
or equivalently:
\begin{align*}
&\frac{\gamma_{0} \tilde{\delta}_{0}}{\sqrt{p}} {\rm Tr}\big[ \boldsymbol{\Sigma}_{i} \mathbf{T}_{1}\left(\boldsymbol{\Sigma}_{0}-\boldsymbol{\Sigma}_{1}\right) \mathbf{T}_{0}\big]\\&+\frac{\gamma_{0} \tilde{\delta}_{0}-\gamma_{1} \tilde{\delta}_{1}}{\sqrt{p}} {\rm Tr}\big[ \boldsymbol{\Sigma}_{i} \mathbf{T}_{1} \boldsymbol{\Sigma}_{1} \mathbf{T}_{0}\big]=\Theta(1)
\end{align*}
Using Assumption 4, it can be readily seen that the first term $\frac{\gamma_{0} \tilde{\delta}_{0}}{\sqrt{p}} {\rm Tr}\big[ \boldsymbol{\Sigma}_{i} \mathbf{T}_{1}\left(\boldsymbol{\Sigma}_{0}-\boldsymbol{\Sigma}_{1}\right) \mathbf{T}_{0}\big]=\Theta(1)$. To satisfy \eqref{eq:nece}, we thus only need to design $\gamma_0$ and $\gamma_1$ such that:
$$
\gamma_{0}  \tilde{\delta}_{0}-\gamma_{1} \tilde{\delta}_{1}=\Theta(1 / \sqrt{p})
$$
or equivalently:
$$
\gamma_{0}+\frac{\gamma_{0} \gamma_{1}}{n_{1}} {\rm Tr} \big[\boldsymbol{\Sigma}_{1} \mathbf{T}_{1}\big]-\gamma_{1}-\frac{\gamma_{0} \gamma_{1}}{n_{0}} {\rm Tr}\big[ \boldsymbol{\Sigma}_{0} \mathbf{T}_{0}\big]=\Theta(1 / \sqrt{p})
$$
Under Assumption 4, 
$$
\frac{1}{n_0}{\rm Tr}\big[ \boldsymbol{\Sigma}_0{\bf T}_0\big]=\frac{1}{n_0}{\rm Tr}\big[\boldsymbol{\Sigma}_1{\bf T}_1\big]+O(\frac{1}{\sqrt{p}})
$$
which proves that in choosing $\gamma_1$ given by:
$$
\gamma_1=\frac{\gamma_{0}}{1-\left(\frac{1}{n_{1}}-\frac{1}{n_{0}}\right) \gamma_{0} {\rm Tr} \big[\boldsymbol{\Sigma}_{0} \mathbf{T}_{0}\big]}
$$
the condition \eqref{eq:nece} becomes satisfied.

\section*{Appendix B.}

The choice of the regularization parameters $\gamma_0$ and $\gamma_1$ allows to ensure that:
$$
\overline{B}_0=\overline{B}_1+O(\frac{1}{\sqrt{p}})
$$
As a result, the expression of the asymptotic equivalents for the classification error rate of both classes   defined in  \eqref{13} for $i \in\{0,1\}$ can be reduced to:
\begin{equation}
\epsilon_{i}^{R-Q D A}-\Phi\left((-1)^{i} \frac{\overline{\xi}_{i}-\overline{b_{i}}}{\sqrt{2 \overline{B_{0}}}}\right) \stackrel{p}{\rightarrow} 0
\end{equation}
Then, the total classification error can be written as:
\begin{align*}
\epsilon^{R-Q D A}&=\pi_{0} \Phi\left(\frac{\beta_{0}+\theta}{\alpha}\right)  +\pi_{1}  \Phi\left(\frac{\beta_{1}-\theta}{\alpha}\right)
\end{align*}
\begin{align*}
    \text{where}    \begin{cases}\beta_{0}=\frac{1}{\sqrt{p}}\left[- \boldsymbol{\mu}^{T} \mathbf{T}_{1} \boldsymbol{\mu}\right]-\frac{1}{\sqrt{p}} {\rm Tr}  \big[\boldsymbol{\Sigma}_{0}\left(\mathbf{T}_{1}-\mathbf{T}_{0}\right) \big]  \\
    \beta_{1}= \frac{1}{\sqrt{p}}\left[- \boldsymbol{\mu}^{T} \mathbf{T}_{0} \boldsymbol{\mu}\right]+\frac{1}{\sqrt{p}} {\rm Tr}\big[ \boldsymbol{\Sigma}_{1}\left(\mathbf{T}_{1}-\mathbf{T}_{0}\right)\big]\\
    \alpha=\sqrt{2 \overline{B_{0}}}
    \end{cases}
\end{align*}
Taking the derivative of this expression with respect to $\theta$ and setting it to zero, the optimal bias $\theta^\star$ should satisfy:
\begin{align*}
\frac{\pi_{0}}{\pi_{1}}e^{(\frac{\beta_{1}-{\theta^\star}}{2\alpha})^{2}-(\frac{\beta_{0}+{\theta^\star}}{2\alpha})^{2}}=1
\end{align*}
Applying the logarithmic function on both sides, we obtain:
\begin{align*}
 \log(\frac{\pi_{0}}{\pi_{1}})+\left(\frac{\beta_{1}-{\theta}^\star}{2\alpha}\right)^{2}-\left(\frac{\beta_{0}+{\theta}^\star}{2\alpha}\right)^{2}=0
\end{align*}
thus leading to
\begin{align*}
 \theta^{*}=\frac{\beta_{1}-\beta_{0}}{2}-\frac{2\alpha^{2}}{\beta_{1}+\beta_{0}}\log(\frac{\pi_{1}}{\pi_{0}})
\end{align*}
\section*{Appendix C}
\hypertarget{Appendix C}{}
In Theorem 4, we provide a consistent estimator for the regularization parameter $\gamma_1$ that satisfies \eqref{eq:necessary} with high probability and a consistent estimator for the optimal bias $\theta^\star$. 
\subsection{Consistent estimator for $\gamma_1$}
We start by proving that $\gamma_1-\hat{\gamma}_1 \stackrel{as}{\rightarrow}    0$. To this end, we need to provide a consistent estimator for ($\frac{1}{n_1}-\frac{1}{n_0}){\rm Tr} \big[\boldsymbol{\Sigma}_0{\bf T}_0\big]$.  We start by noticing that:
$$
(\frac{1}{n_1}-\frac{1}{n_0}){\rm Tr}\big[ \boldsymbol{\Sigma}_0{\bf T}_0\big]=(\frac{n_0}{n_1}-1)\delta_0
$$
A consistent estimator for $\delta_0$ has been provided in \cite{kh} and is given by:
$$
\hat{\delta}_0=\frac{1}{\gamma_0} \frac{\frac{p}{n_0}-\frac{1}{n_0}{\rm Tr}\big[{\bf H}_0(\gamma_0)\big]}{1-\frac{p}{n_0}+\frac{1}{n_0}{\rm Tr}\big[{\bf H}_0(\gamma_0)\big]}
$$
and as such a consistent estimator for $\gamma_1$ in \eqref{eq:gamma} is given by:
$$
\hat{\gamma}_1=\frac{\gamma_0}{1-\gamma_0(\frac{n_0}{n_1}\hat{\delta}_0-\hat{\delta}_0)}
$$
Note that the replacement of $\gamma_1$ by $\hat{\gamma}_1$ still ensures condition \eqref{eq:nece} since from standard results of random matrix theory $\hat{\delta}_0-\delta_0=O(\frac{1}{p})$ with high probability.
\subsection{Consistent estimator for $\theta^\star$}
Recall that 
$$
\theta^\star=\frac{\beta_1-\beta_0}{2}-\frac{2\alpha^2}{\beta_1+\beta_0}\log(\frac{\pi_1}{\pi_0})
$$
To provide a consistent estimator for $\theta^\star$, it is thus required to provide that of $\beta_0,\beta_1$ and $\alpha$. Since $\alpha=\sqrt{2{\overline{B}_0}}$ and $\hat{B}_0-\overline{B}_0\stackrel{\mathrm{a.s.}}{\rightarrow} 0$, we thus have:
$\hat{\alpha}-\alpha \stackrel{\mathrm{a.s.}}{\rightarrow} 0$ where $\hat{\alpha}=\sqrt{2{\hat{B}_0}}$.
As for $\beta_i$, $i=0,1$, it can be written as:
\begin{align*}
	\beta_i&=-\frac{1}{\sqrt{p}} \boldsymbol{\mu}^{T}{\bf T}_{1-i}\boldsymbol{\mu} +\frac{1}{\sqrt{p}}{\rm Tr}\big[\boldsymbol{\Sigma}_i{\bf T}_i\big]-\frac{1}{\sqrt{p}}{\rm Tr}\big[\boldsymbol{\Sigma}_i{\bf T}_{1-i}\big]\\
	&=-\frac{1}{\sqrt{p}} \boldsymbol{\mu}^{T}{\bf T}_{1-i}\boldsymbol{\mu} -\frac{1}{\sqrt{p}}{\rm Tr}\big[\boldsymbol{\Sigma}_i{\bf T}_{1-i}\big]+\frac{n_i}{\sqrt{p}}\delta_i
\end{align*}
Due to the independence of $\boldsymbol{\Sigma}_i$ from ${\bf H}_{1-i}$ and of $\hat{\boldsymbol{\mu}}_1$ and $\hat{\boldsymbol{\mu}}_0$ and ${\bf H}_{i}$, $i=0,1$, we have:
$$
\frac{1}{\sqrt{p}}{\rm Tr} \big[\hat{\boldsymbol{\Sigma}}_i{\bf H}_{1-i}\big]-\frac{1}{\sqrt{p}}{\rm Tr}\big[ {\boldsymbol{\Sigma}}_i{\bf T}_{1-i}\big] \stackrel{as}{\rightarrow}  0
$$
and
$$
\frac{1}{\sqrt{p}}(\hat{\boldsymbol{\mu}}_0-\hat{\boldsymbol{\mu}}_1) {\bf H}_{1-i} (\hat{\boldsymbol{\mu}}_0-\hat{\boldsymbol{\mu}}_1)- \frac{1}{\sqrt{p}}(\hat{\boldsymbol{\mu}}_0-\hat{\boldsymbol{\mu}}_1){\bf T}_{1-i} (\hat{\boldsymbol{\mu}}_0-\hat{\boldsymbol{\mu}}_1) \stackrel{as}{\rightarrow} 0. 
$$
\ifCLASSOPTIONcaptionsoff
  \newpage
\fi



%
\bibliographystyle{IEEEtran}
\bibliography{References}

%








\end{document}